\documentclass[11pt]{article}

\usepackage[final]{acl}

\usepackage{times}
\usepackage{latexsym}

\usepackage[T1]{fontenc}
\usepackage[utf8]{inputenc}
\usepackage{microtype}
\usepackage{inconsolata}

\usepackage{graphicx}
\usepackage{booktabs}
\usepackage{multirow}
\usepackage{array}
\usepackage{amsmath}
\usepackage{amssymb}
\usepackage{placeins}
\newcommand{\dhyp}{\ensuremath{\delta}}
\newcommand{\reldhyp}{\ensuremath{\delta_{\mathrm{rel}}}}

\title{A Hyperbolicity Atlas of Large Language Model Hidden States}

\author{
\textbf{Zhichao Yang$^{1}$, Yuanze Hu$^{1}$, Gen Li$^{1}$, Qingchen Yu$^{1}$, Shiying Duan$^{1}$} \\
\textbf{Xinyu Wang$^{1}$, Ye Qiu$^{1}$, Zeming Liu$^{1}$, Guangxu Chen$^{2}$, Zhaoxin Fan$^{1}$}\thanks{Corresponding author: \texttt{zhaoxinf@buaa.edu.cn}.} \\
$^{1}$Beijing Advanced Innovation Center for Future Blockchain and Privacy Computing,\\
School of Artificial Intelligence, Beihang University \\
$^{2}$South China University of Technology \\
}

\begin{document}
\maketitle

\begin{abstract}
LLM hidden states are ordinary vectors, but the distances among those vectors may still show hierarchical structure.
To our knowledge, this paper is the first systematic study of whether prompt-token hidden states in contemporary LLMs exhibit Gromov Hyperbolicity (GH), a distance-based measure of tree-likeness.
Using 818,904 sample-layer measurements from ten open-weight models across MATH500, HumanEval, WinoGrande, and TruthfulQA, we build a GH map over four axes: parameter scale, layer depth, model family, and input domain.
The clearest pattern is depth, not scale: middle layers usually form a high-\reldhyp{} plateau, while final layers often become substantially more tree-like.
Scale effects are weak and non-monotonic, matched 7/8B model families differ strongly, and domains interact with model specialization.
These findings make GH useful as a practical diagnostic: it shows where hierarchical distance structure appears, how specialization changes it, and which model--layer--domain comparisons deserve closer analysis.
\end{abstract}

\section{Introduction}

Transformer language models store token information in Euclidean vectors, yet many things they process are hierarchical: lexical taxonomies, entailment relations, syntactic trees, discourse dependencies, program blocks, and multi-step derivations.
Hyperbolic geometry is useful for analyzing such structure because negatively curved spaces can represent trees and hierarchies with low distortion \citep{nickel2017poincare,nickel2018learning,ganea2018hyperbolic}.
This observation has already shaped hyperbolic word and sentence embeddings \citep{dhingra2018embedding,tifrea2018poincar}, multilingual and multimodal representation guidelines \citep{sawhney2024adapt}, and recent hyperbolic fine-tuning or architecture proposals for LLMs \citep{yang2026hyperbolic,he2026helm,patil2025hyperbolic}.
For contemporary LLMs, however, an important question remains open: even when the model is not trained in a hyperbolic manifold, do its internal hidden states exhibit tree-like metric organization?

Existing representation analyses answer related but different questions.
Probing classifiers test whether a feature is recoverable; anisotropy and intrinsic-dimension analyses characterize spread or local complexity; structural probes show that some linguistic relations are geometrically recoverable \citep{ethayarajh2019contextual,hewitt2019structural,valeriani2023geometry}.
These methods do not directly test whether distances among hidden states behave like a hierarchy.
Gromov Hyperbolicity (GH) provides a simple complementary test: it asks whether distances among hidden-state vectors behave more like distances in a tree.
To our knowledge, this paper is the first systematic study of GH for prompt-token hidden states in modern open-weight LLMs.

We use the word ``atlas'' in a literal sense: we map the statistic across several axes instead of giving one number per model.
For a model $m$, layer $\ell$, domain $d$, and prompt $x$, we measure the normalized statistic
\begin{equation}
\mathcal{A}(m,\ell,d,x) = \reldhyp\!\left(X_{m,\ell,x}\right),
\label{eq:atlas-map}
\end{equation}
where $X_{m,\ell,x}$ is the set of prompt-token hidden states for that layer.
This map is necessary because the axes need not move together.
A larger model may change the final layer without changing the middle layers; a domain may be tree-like for one specialization but not for another; and a family difference at matched parameter count can exceed a scale difference within a family.
The empirical question is therefore not whether ``LLMs are hyperbolic,'' but where lower or higher GH appears across models, layers, and domains.

We organize the study around four questions: 1) within a family, does normalized hyperbolicity change when scale grows from 1.5B to 7B parameters? 2) Within a model, how does hyperbolicity vary across transformer layers? 3) At matched scale, do comparable 7/8B model families exhibit different final-layer profiles? 4) For a fixed model, how does hyperbolicity change across math, code, commonsense, and truthfulness prompts?

To answer these questions, we compute prompt-only hidden states for ten open-weight checkpoints from Llama, Qwen2.5, Qwen2.5-Math, Qwen2.5-Coder, and DeepSeek-R1-Distill-Qwen families on MATH500, HumanEval, WinoGrande, and TruthfulQA.
For each prompt and layer, we sample quadruples of token hidden states, estimate four-point GH, and normalize by the largest pairwise distance in that set.
This design keeps the comparisons aligned: within-family scale contrasts, layerwise trajectories, matched 7/8B family comparisons, and fixed-model domain comparisons are all produced by the same measurement protocol.
It also avoids a common shortcut: generated responses are excluded so that prompt geometry is not confounded with decoding length, answer correctness, or model-specific generation behavior.

The resulting map yields four conclusions.
First, scale alone is weak and non-monotonic: Qwen2.5-7B-Base is more tree-like than Qwen2.5-1.5B on some domains but less tree-like on others, while the math track moves upward in \reldhyp{}.
Second, depth dominates the geometry: middle layers form a high-\reldhyp{} plateau, and final layers usually drop toward more tree-like structure.
Third, matched 7/8B families occupy distinct value ranges, with family and specialization gaps larger than the pooled scale shifts.
Fourth, domains interact with specialization: HumanEval is high-\reldhyp{} for Qwen2.5-7B-Base but becomes the lowest-\reldhyp{} domain for Qwen2.5-7B-Coder-Instruct.

Our contribution is this GH map and its interpretation as a diagnostic tool.
GH does not prove symbolic reasoning or identify a causal circuit by itself.
Its value is narrower and more concrete: it tells us where prompt-token distances become more tree-like, where they do not, and which model--layer--domain comparisons deserve closer follow-up.

\section{Background}

\subsection{Hyperbolicity as Tree-Likeness}

Gromov hyperbolicity measures how close a metric space is to a tree metric \citep{gromov1987hyperbolic}.
For four points $a,b,c,d$ with pairwise distance $d(\cdot,\cdot)$, define
\begin{align}
s_1 &= d(a,b) + d(c,d), \\
s_2 &= d(a,c) + d(b,d), \\
s_3 &= d(a,d) + d(b,c).
\end{align}
Let $s_{(1)} \leq s_{(2)} \leq s_{(3)}$ be the sorted values.
The four-point hyperbolicity of the quadruple is
\begin{equation}
\dhyp(a,b,c,d) = \frac{1}{2}\left(s_{(3)} - s_{(2)}\right).
\label{eq:delta}
\end{equation}
A tree metric has $\dhyp=0$ for every quadruple.
Lower values therefore indicate a more tree-like metric structure, while larger values indicate stronger deviation from tree-likeness.

Raw $\dhyp$ depends on the size of the distances, so it is hard to compare models with different hidden dimensions and activation norms directly.
We report a diameter-normalized statistic
\begin{equation}
\reldhyp(X) = \frac{\widehat{\dhyp}(X)}{\operatorname{diam}(X)},
\label{eq:relative-delta}
\end{equation}
where $X$ is a set of hidden-state vectors, $\widehat{\dhyp}$ is the sampled four-point estimate, and $\operatorname{diam}(X)$ is the maximum observed pairwise distance in the sampled set.
In the rest of the paper, lower \reldhyp{} means that the token-state distance pattern is more tree-like after normalization.

\subsection{Prompt Hidden-State Sets}

For model $m$, layer $\ell$, prompt $x$, and token position $t$, let $h_{m,\ell}(x,t) \in \mathbb{R}^{d_m}$ denote the hidden state.
For each prompt and layer, we collect the hidden states of all prompt tokens:
\begin{equation}
X_{m,\ell,x} = \{h_{m,\ell}(x,t): t \in T(x)\},
\end{equation}
where $T(x)$ is the set of prompt-token positions.
We use Euclidean distances over the model hidden-state coordinates and estimate \reldhyp{} separately for each sample and layer.
This prompt-only design isolates geometry induced by the input distribution from geometry induced by model-specific generated text length, decoding behavior, and answer correctness.

\section{Experimental Design}

\subsection{Models, Data, and Scope}

We analyze ten open-weight checkpoints from the Llama, Qwen2.5, Qwen2.5-Math, Qwen2.5-Coder, and DeepSeek-R1-Distill-Qwen families \citep{grattafiori2024llama,yang2024qwen2,hui2024qwen2,guo2025deepseek}.
The 1.5B models include Qwen2.5-1.5B, Qwen2.5-1.5B-Instruct, Qwen2.5-Math-1.5B, and DeepSeek-R1-Distill-Qwen-1.5B.
The 7/8B set includes Llama-3.1-8B, Llama-3-8B-Instruct, Qwen2.5-7B-Base, Qwen2.5-7B-Instruct, Qwen2.5-7B-Coder-Instruct, and Qwen2.5-Math-7B.

The datasets cover mathematical reasoning (MATH500, derived from MATH and verifier work), code generation (HumanEval), commonsense pronoun resolution (WinoGrande), and truthfulness (TruthfulQA) \citep{hendrycks2021measuring,lightman2024let,chen2021evaluating,sakaguchi2021winogrande,lin2022truthfulqa}.
The sample counts are 500 for MATH500, 164 for HumanEval, 1,267 for WinoGrande, and 817 for TruthfulQA for each model, yielding the same dataset composition across checkpoints.
The measurements are prompt-only: generated responses in the local result directory are not used for the main atlas because generation length and answer correctness would add another uncontrolled axis.

\subsection{Measurement Protocol}

For every model, dataset example, and extracted layer, we compute prompt-token hidden states without gradient updates.
One measurement uses the prompt-token hidden states from a single example at a single layer.
For that set, we sample 2,000 quadruples of token states and compute the normalized four-point estimate in Eq.~\ref{eq:relative-delta}.
We retain the embedding layer and every available transformer layer, giving 818,904 valid sample-layer measurements.

The main analysis uses medians and interquartile ranges because per-example \reldhyp{} distributions are skewed and often bounded near 1 in middle layers.
For paired layer comparisons we use the Wilcoxon signed-rank test over matched examples.
For unpaired family and domain comparisons we use Kruskal--Wallis tests and report median shifts or ranges.
We interpret the results as distributions, not as yes-or-no labels.
A lower median does not prove that a model stores an explicit symbolic tree; it means the sampled token-state distances are closer to a tree metric under the four-point test.

\section{Results}

Lower \reldhyp{} means the prompt-token distance pattern is closer to a tree metric after diameter normalization.
All tables report medians with interquartile ranges unless otherwise stated.

\subsection{RQ1: Scale Effects Are Conditional}

\begin{figure*}[t]
\centering
\includegraphics[width=\textwidth]{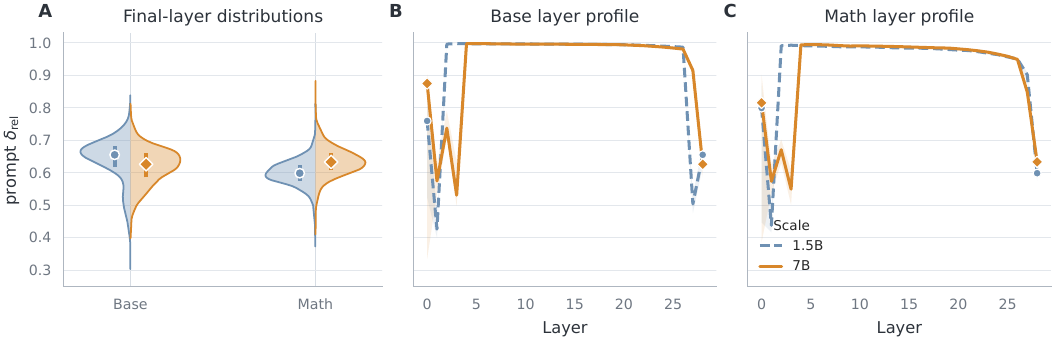}
\caption{RQ1 scale comparison within the Qwen2.5 base and Qwen2.5-Math tracks. Panel A shows final-layer distributions pooled over domains; panels B--C show median layer profiles.}
\label{fig:rq1-scale-composite}
\end{figure*}

\begin{table*}[t]
\centering
\small
\setlength{\tabcolsep}{4pt}
\renewcommand{\arraystretch}{1.02}
\begin{tabular*}{0.80\textwidth}{@{\extracolsep{\fill}}llccc@{}}
\toprule
\textbf{Track} & \textbf{Dataset} & \textbf{1.5B} & \textbf{7B} & \textbf{7B--1.5B} \\ 
\midrule
Base & MATH500 & \shortstack{\textbf{0.526}\\[-2.0pt]{\scriptsize\color{gray}[0.486, 0.568]}} & \shortstack{0.549\\[-2.0pt]{\scriptsize\color{gray}[0.519, 0.591]}} & +0.024 \\
 & HumanEval & \shortstack{\textbf{0.655}\\[-2.0pt]{\scriptsize\color{gray}[0.636, 0.687]}} & \shortstack{0.676\\[-2.0pt]{\scriptsize\color{gray}[0.640, 0.712]}} & +0.021 \\
 & WinoGrande & \shortstack{0.670\\[-2.0pt]{\scriptsize\color{gray}[0.649, 0.694]}} & \shortstack{\textbf{0.648}\\[-2.0pt]{\scriptsize\color{gray}[0.624, 0.670]}} & -0.022 \\
 & TruthfulQA & \shortstack{0.655\\[-2.0pt]{\scriptsize\color{gray}[0.628, 0.680]}} & \shortstack{\textbf{0.604}\\[-2.0pt]{\scriptsize\color{gray}[0.573, 0.633]}} & -0.051 \\
\addlinespace[2pt]
\midrule
Math & MATH500 & \shortstack{\textbf{0.581}\\[-2.0pt]{\scriptsize\color{gray}[0.539, 0.621]}} & \shortstack{0.600\\[-2.0pt]{\scriptsize\color{gray}[0.566, 0.630]}} & +0.019 \\
 & HumanEval & \shortstack{\textbf{0.600}\\[-2.0pt]{\scriptsize\color{gray}[0.572, 0.634]}} & \shortstack{0.694\\[-2.0pt]{\scriptsize\color{gray}[0.670, 0.725]}} & +0.093 \\
 & WinoGrande & \shortstack{\textbf{0.604}\\[-2.0pt]{\scriptsize\color{gray}[0.585, 0.627]}} & \shortstack{0.644\\[-2.0pt]{\scriptsize\color{gray}[0.620, 0.667]}} & +0.039 \\
 & TruthfulQA & \shortstack{\textbf{0.593}\\[-2.0pt]{\scriptsize\color{gray}[0.573, 0.619]}} & \shortstack{0.624\\[-2.0pt]{\scriptsize\color{gray}[0.605, 0.646]}} & +0.031 \\
\bottomrule
\end{tabular*}
\caption{RQ1 within-family scale comparison by domain. Entries are final-layer medians with interquartile ranges; the last column is the 7B minus 1.5B median shift.}
\label{tab:rq1-scale-summary}
\end{table*}

Figure~\ref{fig:rq1-scale-composite} and Table~\ref{tab:rq1-scale-summary} show that parameter scale alone is a weak predictor of prompt geometry.
Pooled over domains, the base track shifts by only -0.029 from 1.5B to 7B, while the math track shifts by +0.035; the pooled Cliff's $\Delta$ is +0.08.
The small aggregate effect hides domain-specific reversals.
For the base track, the larger model is less tree-like on MATH500 and HumanEval (+0.024 and +0.021), but more tree-like on WinoGrande and TruthfulQA (-0.022 and -0.051).
For the math track, the 7B model is higher-\reldhyp{} in every domain, with the largest shift on HumanEval (+0.093), suggesting that more math-specialized capacity does not automatically produce more tree-like final prompt geometry.

The direction of the base-track shift is especially informative.
MATH500 and HumanEval are the two domains with the most explicit formal structure, yet they do not both become more tree-like with scale; in fact, both move upward in \reldhyp{} for Qwen2.5-7B-Base.
Conversely, WinoGrande and TruthfulQA move downward, even though they are less obviously tree-structured tasks.
This indicates that the statistic is not simply tracking whether a dataset name sounds hierarchical.
It is tracking how a particular family reorganizes prompt-token distances after training and scaling.
The weak pooled effect is therefore informative: scale matters less than the combination of family, domain, and layer.

The layer profiles explain why final-layer scale summaries should be read cautiously.
Both tracks quickly enter a middle-layer plateau near \reldhyp{}$\approx 1$, and most of the visible scale effect occurs in the first few layers and in the final exit from the plateau.
Thus scaling does not translate the entire curve upward or downward.
It changes when the model enters the high-\reldhyp{} middle-layer range and how sharply it leaves that range; both changes depend on family and domain.

\paragraph{Takeaway.}
Parameter count alone is not a reliable explanation for GH.
Scale should be reported together with model family, domain, and layer.

\subsection{RQ2: Depth Dominates the Atlas}

\begin{figure*}[t]
\centering
\includegraphics[width=\textwidth]{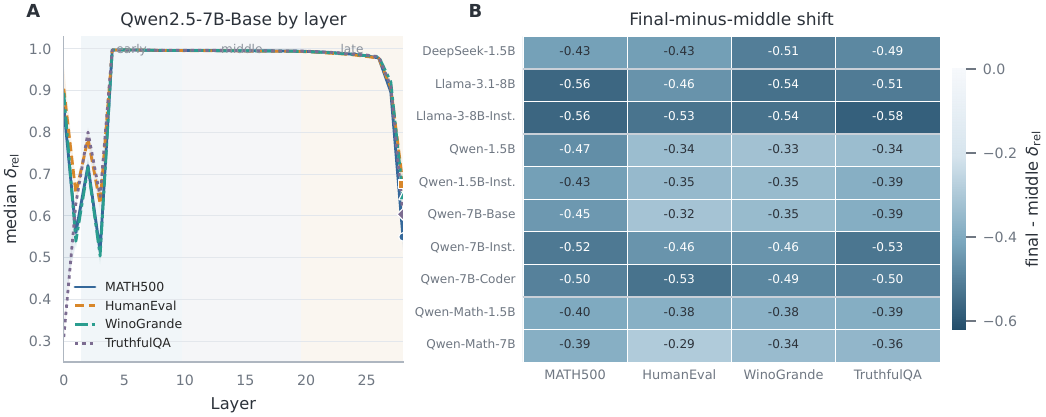}
\caption{RQ2 depth effects. Panel A shows the layer trajectory for Qwen2.5-7B-Base. Panel B shows the median final-minus-middle shift for every model--domain pair. Negative values mean the final layer is more tree-like than the middle-layer median.}
\label{fig:rq2-depth-composite}
\end{figure*}

\begin{table*}[t]
\centering
\small
\setlength{\tabcolsep}{3pt}
\renewcommand{\arraystretch}{1.02}
\begin{tabular*}{0.82\textwidth}{@{\extracolsep{\fill}}lcccc@{}}
\toprule
\textbf{Dataset} & \textbf{Embedding} & \textbf{Middle} & \textbf{Final} & \textbf{Final--Middle} \\ 
\midrule
MATH500 & \shortstack{0.860\\[-2.0pt]{\scriptsize\color{gray}[0.384, 0.908]}} & \shortstack{0.996\\[-2.0pt]{\scriptsize\color{gray}[0.995, 0.996]}} & \shortstack{0.549\\[-2.0pt]{\scriptsize\color{gray}[0.519, 0.591]}} & -0.446 \\
HumanEval & \shortstack{0.905\\[-2.0pt]{\scriptsize\color{gray}[0.880, 0.962]}} & \shortstack{0.995\\[-2.0pt]{\scriptsize\color{gray}[0.995, 0.995]}} & \shortstack{0.676\\[-2.0pt]{\scriptsize\color{gray}[0.640, 0.712]}} & -0.319 \\
WinoGrande & \shortstack{0.895\\[-2.0pt]{\scriptsize\color{gray}[0.876, 0.911]}} & \shortstack{0.995\\[-2.0pt]{\scriptsize\color{gray}[0.995, 0.996]}} & \shortstack{0.648\\[-2.0pt]{\scriptsize\color{gray}[0.624, 0.670]}} & -0.347 \\
TruthfulQA & \shortstack{0.310\\[-2.0pt]{\scriptsize\color{gray}[0.296, 0.339]}} & \shortstack{0.996\\[-2.0pt]{\scriptsize\color{gray}[0.996, 0.996]}} & \shortstack{0.604\\[-2.0pt]{\scriptsize\color{gray}[0.573, 0.633]}} & -0.392 \\
\bottomrule
\end{tabular*}
\caption{RQ2 depth summary for Qwen2.5-7B-Base. Middle is the per-example median over relative layer positions 0.35--0.70.}
\label{tab:rq2-depth-summary}
\end{table*}

Layer depth is the most stable factor in the study.
For Qwen2.5-7B-Base, Figure~\ref{fig:rq2-depth-composite}A shows a three-part trajectory: early layers depend on the input, middle layers form a long high-\reldhyp{} plateau, and final layers drop sharply.
Table~\ref{tab:rq2-depth-summary} quantifies the same pattern.
The middle median is almost identical across domains (0.995--0.996), while final medians spread from 0.549 on MATH500 to 0.676 on HumanEval.
The final-minus-middle shift is negative for every domain, ranging from -0.319 on HumanEval to -0.446 on MATH500.

The embedding layer shows why the final drop is not just inherited from the input representation.
MATH500, HumanEval, and WinoGrande begin with relatively high embedding-layer medians, whereas TruthfulQA begins much lower (0.310) and then rises to 0.604 at the final layer.
In other words, the network can either sharpen tree-like structure relative to the embedding layer or weaken it, while still ending far below the middle-layer plateau.
Across Figure~\ref{fig:rq2-depth-composite}B, every model--domain pair has a negative final-minus-middle shift, but the magnitude varies strongly.
Llama and Qwen-Coder variants often show drops near -0.5, whereas Qwen2.5-Math-7B on HumanEval drops by only -0.29.
Depth is therefore a robust effect, but the size of the final-layer drop still depends on the model and the domain.

This result also clarifies the role of the middle-layer plateau.
Because the plateau appears across domains with very different prompt lengths and lexical distributions, it is unlikely to be a narrow artifact of one dataset.
At the same time, the plateau does not make the statistic useless: final layers consistently leave the plateau and separate models and domains again.
The simplest interpretation is computational rather than architectural.
Middle layers maintain many competing relational constraints at once, making the resulting distance pattern less compatible with a single tree metric.
Final layers then compress the representation toward the information needed by the next-token objective, and that compression can produce a more tree-like distance structure.

\paragraph{Takeaway.}
The main geometric change happens across depth: most models move from a high-\reldhyp{} middle layer to a lower-\reldhyp{} final layer.

\subsection{RQ3: Matched Families Occupy Different Regimes}

\begin{figure*}[t]
\centering
\includegraphics[width=\textwidth]{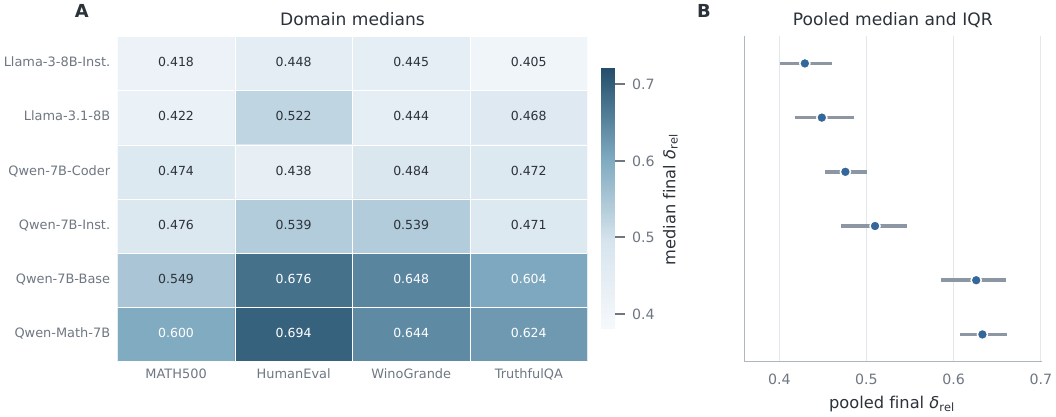}
\caption{RQ3 matched 7/8B family comparison. Panel A reports final-layer domain medians; panel B summarizes the pooled final-layer median and IQR for each model.}
\label{fig:rq3-family-composite}
\end{figure*}

\begin{table*}[t]
\centering
\small
\setlength{\tabcolsep}{3.5pt}
\renewcommand{\arraystretch}{1.02}
\begin{tabular*}{0.90\textwidth}{@{\extracolsep{\fill}}p{0.24\textwidth}cccc@{}}
\toprule
\textbf{Model} & \textbf{MATH500} & \textbf{HumanEval} & \textbf{WinoGrande} & \textbf{TruthfulQA} \\ 
\midrule
Llama-3.1-8B & \shortstack{\underline{0.422}\\[-2.0pt]{\scriptsize\color{gray}[0.373, 0.482]}} & \shortstack{0.522\\[-2.0pt]{\scriptsize\color{gray}[0.435, 0.590]}} & \shortstack{\textbf{0.444}\\[-2.0pt]{\scriptsize\color{gray}[0.418, 0.471]}} & \shortstack{\underline{0.468}\\[-2.0pt]{\scriptsize\color{gray}[0.435, 0.493]}} \\
Llama-3-8B-Instruct & \shortstack{\textbf{0.418}\\[-2.0pt]{\scriptsize\color{gray}[0.385, 0.473]}} & \shortstack{\underline{0.448}\\[-2.0pt]{\scriptsize\color{gray}[0.388, 0.556]}} & \shortstack{\underline{0.445}\\[-2.0pt]{\scriptsize\color{gray}[0.423, 0.468]}} & \shortstack{\textbf{0.405}\\[-2.0pt]{\scriptsize\color{gray}[0.377, 0.430]}} \\
\addlinespace[2pt]
Qwen2.5-7B-Base & \shortstack{0.549\\[-2.0pt]{\scriptsize\color{gray}[0.519, 0.591]}} & \shortstack{0.676\\[-2.0pt]{\scriptsize\color{gray}[0.640, 0.712]}} & \shortstack{0.648\\[-2.0pt]{\scriptsize\color{gray}[0.624, 0.670]}} & \shortstack{0.604\\[-2.0pt]{\scriptsize\color{gray}[0.573, 0.633]}} \\
Qwen2.5-7B-Instruct & \shortstack{0.476\\[-2.0pt]{\scriptsize\color{gray}[0.448, 0.500]}} & \shortstack{0.539\\[-2.0pt]{\scriptsize\color{gray}[0.506, 0.574]}} & \shortstack{0.539\\[-2.0pt]{\scriptsize\color{gray}[0.514, 0.568]}} & \shortstack{0.471\\[-2.0pt]{\scriptsize\color{gray}[0.441, 0.502]}} \\
Qwen2.5-7B-Coder-Instruct & \shortstack{0.474\\[-2.0pt]{\scriptsize\color{gray}[0.444, 0.500]}} & \shortstack{\textbf{0.438}\\[-2.0pt]{\scriptsize\color{gray}[0.417, 0.464]}} & \shortstack{0.484\\[-2.0pt]{\scriptsize\color{gray}[0.456, 0.515]}} & \shortstack{0.472\\[-2.0pt]{\scriptsize\color{gray}[0.457, 0.490]}} \\
Qwen2.5-Math-7B & \shortstack{0.600\\[-2.0pt]{\scriptsize\color{gray}[0.566, 0.630]}} & \shortstack{0.694\\[-2.0pt]{\scriptsize\color{gray}[0.670, 0.725]}} & \shortstack{0.644\\[-2.0pt]{\scriptsize\color{gray}[0.620, 0.667]}} & \shortstack{0.624\\[-2.0pt]{\scriptsize\color{gray}[0.605, 0.646]}} \\
\bottomrule
\end{tabular*}
\caption{RQ3 matched-scale family summary. Each cell reports final-layer median $\delta_{\mathrm{rel}}$ with interquartile range. Lower values indicate more tree-like metric structure. Bold and underlined medians are the lowest and second-lowest in each column.}
\label{tab:curvature-summary}
\end{table*}

At matched scale, family differences are larger than the within-family scale effects in RQ1.
Figure~\ref{fig:rq3-family-composite} and Table~\ref{tab:curvature-summary} show that pooled matched-model medians range from 0.429 for Llama-3-8B-Instruct to 0.633 for Qwen2.5-Math-7B, with a Kruskal--Wallis test giving $p<10^{-300}$.
The cell-level spread is also substantial: TruthfulQA on Llama-3-8B-Instruct is 0.405, while HumanEval on Qwen2.5-Math-7B is 0.694.
This range is much larger than the pooled scale shifts of -0.029 and +0.035.

The matched-family ordering is not merely ``Llama versus Qwen.''
Llama models are consistently low-\reldhyp{}, but Qwen variants separate according to specialization.
Qwen2.5-7B-Base and Qwen2.5-Math-7B have the highest final-layer values, especially on HumanEval and WinoGrande.
Qwen2.5-7B-Instruct lowers \reldhyp{} across all domains, and Qwen2.5-7B-Coder-Instruct is closer to the Llama models than to Qwen2.5-7B-Base.
Architecture, tokenizer, pretraining data, and post-training are mixed together here, so we should not assign the gap to one cause.
The practical point is clear: models at the same approximate scale can organize prompt hidden states in measurably different ways.

The within-domain ordering strengthens this conclusion.
On HumanEval, the gap between Qwen2.5-Math-7B and Qwen2.5-7B-Coder-Instruct is 0.256, far larger than any within-family scale shift in Table~\ref{tab:rq1-scale-summary}.
On TruthfulQA, Llama-3-8B-Instruct reaches 0.405 while Qwen2.5-Math-7B remains at 0.624, again showing that matched-size families can end in very different final-layer geometries.
The repeated pattern is that post-training and specialization can move a model into a different geometric range without changing its nominal scale class.

\paragraph{Takeaway.}
Matched-size models should not be treated as interchangeable in hidden-state geometry.
Family and specialization can matter more than parameter count.

\subsection{RQ4: Domains Interact with Specialization}

\begin{figure*}[t]
\centering
\includegraphics[width=\textwidth]{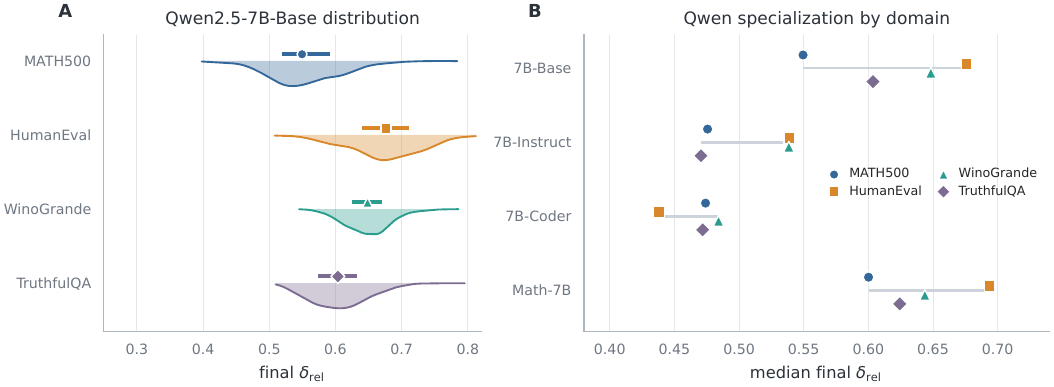}
\caption{RQ4 domain effects. Panel A shows final-layer distributions for Qwen2.5-7B-Base. Panel B compares domain medians across four matched Qwen 7B variants.}
\label{fig:rq4-domain-composite}
\end{figure*}

\begin{table*}[t]
\centering
\small
\setlength{\tabcolsep}{4pt}
\renewcommand{\arraystretch}{1.02}
\begin{tabular*}{0.84\textwidth}{@{\extracolsep{\fill}}lcccc@{}}
\toprule
\textbf{Dataset} & \textbf{7B-Base} & \textbf{7B-Instruct} & \textbf{7B-Coder} & \textbf{Math-7B} \\ 
\midrule
MATH500 & \shortstack{0.549\\[-2.0pt]{\scriptsize\color{gray}[0.519, 0.591]}} & \shortstack{\underline{0.476}\\[-2.0pt]{\scriptsize\color{gray}[0.448, 0.500]}} & \shortstack{\textbf{0.474}\\[-2.0pt]{\scriptsize\color{gray}[0.444, 0.500]}} & \shortstack{0.600\\[-2.0pt]{\scriptsize\color{gray}[0.566, 0.630]}} \\
HumanEval & \shortstack{0.676\\[-2.0pt]{\scriptsize\color{gray}[0.640, 0.712]}} & \shortstack{\underline{0.539}\\[-2.0pt]{\scriptsize\color{gray}[0.506, 0.574]}} & \shortstack{\textbf{0.438}\\[-2.0pt]{\scriptsize\color{gray}[0.417, 0.464]}} & \shortstack{0.694\\[-2.0pt]{\scriptsize\color{gray}[0.670, 0.725]}} \\
WinoGrande & \shortstack{0.648\\[-2.0pt]{\scriptsize\color{gray}[0.624, 0.670]}} & \shortstack{\underline{0.539}\\[-2.0pt]{\scriptsize\color{gray}[0.514, 0.568]}} & \shortstack{\textbf{0.484}\\[-2.0pt]{\scriptsize\color{gray}[0.456, 0.515]}} & \shortstack{0.644\\[-2.0pt]{\scriptsize\color{gray}[0.620, 0.667]}} \\
TruthfulQA & \shortstack{0.604\\[-2.0pt]{\scriptsize\color{gray}[0.573, 0.633]}} & \shortstack{\textbf{0.471}\\[-2.0pt]{\scriptsize\color{gray}[0.441, 0.502]}} & \shortstack{\underline{0.472}\\[-2.0pt]{\scriptsize\color{gray}[0.457, 0.490]}} & \shortstack{0.624\\[-2.0pt]{\scriptsize\color{gray}[0.605, 0.646]}} \\
\bottomrule
\end{tabular*}
\caption{RQ4 domain summary for matched Qwen 7B variants. Each cell reports final-layer median $\delta_{\mathrm{rel}}$ with interquartile range. Bold and underlined medians are the lowest and second-lowest in each row.}
\label{tab:rq4-domain-summary}
\end{table*}

Domain effects are visible even when the model is fixed.
For Qwen2.5-7B-Base, Figure~\ref{fig:rq4-domain-composite}A and Table~\ref{tab:rq4-domain-summary} show a clear ordering: MATH500 is most tree-like (0.549), TruthfulQA is intermediate (0.604), WinoGrande is higher (0.648), and HumanEval is least tree-like (0.676).
This ordering is not explained by sample size alone, because the HumanEval distribution remains high despite being the smallest dataset.

Specialization changes the domain ordering rather than shifting all domains by a constant amount.
Instruction tuning lowers Qwen2.5-7B-Base fairly broadly, for example from 0.676 to 0.539 on HumanEval and from 0.604 to 0.471 on TruthfulQA.
Coder specialization produces the sharpest interaction: HumanEval becomes the lowest-\reldhyp{} domain for Qwen2.5-7B-Coder-Instruct (0.438), a -0.238 shift from the base model.
By contrast, Qwen2.5-Math-7B is not more tree-like than the base model on MATH500; its MATH500 median is 0.600 versus 0.549 for Qwen2.5-7B-Base.
Thus a domain label and a model specialization label do not map one-to-one onto lower GH.
The strongest regularity is relational: code prompts become tree-like in the coder model, while math specialization keeps a higher-\reldhyp{} final geometry across domains.

The domain effect is also visible in the full distributions.
For Qwen2.5-7B-Base, HumanEval is not merely shifted upward by a few outliers; its quartiles sit above the MATH500 distribution, as shown in Figure~\ref{fig:rq4-domain-composite}A.
For Qwen2.5-7B-Coder-Instruct, the same domain becomes the lowest-median condition, which means the reversal is not produced by dataset size or by a fixed property of HumanEval prompts.
Instead, code specialization appears to make code-prompt final-layer distances more tree-like.
This fits the intuition that code has nested structure, but the comparison with Qwen2.5-Math-7B shows that this structure appears clearly only in the relevant specialized model.

Prompt length is a possible confound because longer prompts provide more token states in each sample.
We therefore bin examples by prompt-length quartile within each model and dataset as a control analysis.
For Qwen2.5-7B-Base, HumanEval and WinoGrande remain higher than MATH500 across quartiles, and the within-domain slopes are modest.
For Llama-3-8B-Instruct, medians are lower overall and still do not collapse into a single length-driven trend.
Length can add local variation, but it does not explain the large family and specialization effects above.

\paragraph{Takeaway.}
Domain effects are real, but they depend on the model.
The same dataset can look less tree-like in one model and more tree-like in another.

\subsection{Cross-RQ Patterns}

Taken together, the four RQs reveal three regularities.
First, hyperbolicity is primarily a trajectory rather than an endpoint.
The final layer is important because it is closest to the prediction interface, but the most repeatable phenomenon is the movement from early variability to a high-\reldhyp{} middle plateau and then to a model-specific final drop.
This is why scale and family comparisons are most interpretable when paired with layer profiles.

Second, specialization changes which domains receive a tree-like final organization.
The coder model does not simply lower all medians equally; it selectively makes HumanEval the lowest-\reldhyp{} domain among the Qwen variants.
Instruction tuning produces a broader reduction, while math specialization keeps relatively high final-layer \reldhyp{} even on MATH500.
This pattern suggests that lower hyperbolicity is not a generic marker of task difficulty or formal structure.
It is better understood as an alignment between a model's learned representational organization and the prompt distribution.

Third, family differences at matched scale dominate simple parameter-count explanations.
The Llama models and Qwen-Coder have lower final-layer values than Qwen2.5-7B-Base and Qwen2.5-Math-7B, even though all are in the 7/8B range.

\FloatBarrier

\section{Discussion}

\subsection{Interpreting the Geometric Patterns}

The most robust pattern is the three-part layer trajectory.
Embedding and early layers are sensitive to domain and tokenization.
Middle layers are close to the upper end of the normalized statistic, suggesting a highly non-tree-like organization under the four-point metric.
Final layers then often move toward lower \reldhyp{}, but the degree of movement depends on family and domain.
This supports a depth-centered view: hyperbolicity is not a fixed property of a model, but a trajectory through the computation.

One interpretation is that middle layers mix multiple incompatible organizational principles.
They must preserve lexical identity, local syntax, long-range context, and task cues before the model has fully compressed the prompt into a prediction-oriented representation.
A collection of token states that simultaneously reflects many such axes may be less tree-like than either the input embedding space or the final task-facing representation.
The final-layer drop can then be read as compression toward a simpler distance pattern, although the present measurements do not identify the mechanism causally.

The Qwen variants show that specialization is not merely a performance label.
Coder specialization strongly lowers HumanEval \reldhyp{}, while math specialization does not make every domain more tree-like and can even raise final-layer \reldhyp{} relative to the base track.
This suggests that training or adaptation changes which domains receive a tree-like final organization.
The same domain can therefore have simple distances for one model and more complex distances for another.

\subsection{Scope and Meaning of the Diagnostic}

Here, a low \reldhyp{} means that sampled distances among prompt-token hidden states are closer to a tree metric.
The statistic is nevertheless useful because it depends only on distances, is cheap to summarize as a distribution, and is comparable across models.
It complements probing, intrinsic dimension, anisotropy, and intervention-based analyses \citep{luan2026lyapunov,qin2026achilles}.

The GH map is most informative when it is used comparatively.
A low final-layer value for one model says little about task performance; systematic reversals across specializations can reveal how training changes the geometry assigned to a domain.
For example, HumanEval is high-\reldhyp{} for Qwen2.5-7B-Base and Qwen2.5-Math-7B, but low-\reldhyp{} for Qwen2.5-7B-Coder-Instruct.
The relevant comparison is between models: code prompts become tree-like in the model whose training history makes code structure central.

Geometrically, the middle-layer plateau is a stable part of the layer trajectory: token distances are far from a single tree metric there, and the layers separate again near the output.

To directly address RQ1.2, we examine the same contrast during generation in a controlled HumanEval analysis.
We use 50 fixed HumanEval prompts and Qwen2.5-7B-Base and Qwen2.5-7B-Coder-Instruct.
Each model generates greedily with a budget of 512 new tokens.
We then replay the exact prompt plus generated continuation with teacher forcing, so that hidden states are measured on the model's actual generated text.
The same score is computed separately for prompt tokens, generated-code tokens, and the full sequence, using 1,000 sampled quadruples per example and layer.
At the final layer, the Base/Coder medians are 0.665/0.422 for prompt-only states, 0.676/0.423 for generated-code states, and 0.656/0.418 for the full sequence, giving Coder--Base differences of $-0.242$, $-0.254$, and $-0.238$, respectively.

The contrast appears in 50/50 prompt sets, 45/50 generated-code sets, and 48/50 full sequences, with the Coder model lower in each case.
Coder also generates longer continuations: 208 tokens at the median versus 142 for Base, an increase of 59 tokens.
The generated-code result is also stable under cosine ($-0.356$), token-wise L2-normalized Euclidean ($-0.308$), and centered-L2 ($-0.287$) settings.
Taken together, these results extend the original observation to this fixed generation setting: the HumanEval Base/Coder contrast appears in prompt states and in hidden states for generated code under greedy decoding.
The contrast remains visible after the model's generated continuation is included.

\subsection{Implications for Future Analyses}

These results suggest two practical reporting norms for geometric studies of LLM hidden states.
First, final-layer summaries should be accompanied by at least a coarse depth profile, because middle and final layers answer different questions.
Second, model comparisons should be matched by family, scale, and domain whenever possible.
Otherwise, a domain effect can be mistaken for specialization, and a specialization effect can be mistaken for scale.
Hyperbolicity is therefore better used as one coordinate in a larger representation-analysis suite than as a standalone score.

\section{Related Work}

Hyperbolic representation learning represents hierarchies with low distortion \citep{nickel2017poincare,nickel2018learning} and defines neural operations on hyperbolic manifolds \citep{ganea2018hyperbolic}.
NLP applications include word and sentence embeddings \citep{dhingra2018embedding,tifrea2018poincar} and multilingual or multimodal guidance and alignment \citep{sawhney2024adapt,hu2026tinyalign}.
Representation-geometry studies report anisotropy and layer dependence \citep{ethayarajh2019contextual}, syntactic geometry \citep{hewitt2019structural}, and intrinsic-dimension or neighborhood variation across layers and domains \citep{valeriani2023geometry}; related manifold-connectivity work links representation topology to hallucination mitigation \citep{hu2026cose}.
Recent LLM work studies hyperbolic embeddings and adaptation \citep{yang2026hyperbolic}, mixture-of-curvature models \citep{he2026helm}, broader methods \citep{patil2025hyperbolic}, and adversarial-prompt detection \citep{yung2025geometry}. Complementary diagnostics use dynamical stability probes \citep{luan2026lyapunov} and perturbation-based causal identification of critical neurons \citep{qin2026achilles}.
Most of these works either train a hyperbolic representation, adapt a model, or use geometry for a specific detection task.
Our setting is different: we do not change the model.
We measure the same GH statistic in existing LLM prompt hidden states across scale, depth, family, and data domain.

\section{Conclusion}

We presented a prompt-only hyperbolicity atlas for LLM hidden states across ten models, four datasets, and all layers.
Normalized Gromov hyperbolicity is highly structured: depth dominates, matched families differ, domains interact with specialization, and scale alone is weak and non-monotonic.
Hyperbolicity is therefore a useful diagnostic when interpreted as a model--layer--domain property rather than as a direct claim about symbolic hierarchy.

\section*{Limitations}

The main atlas measures prompt-token hidden states only.
The random-initialization control covers one architecture, one initialization, and two domains.
It provides initial evidence that the trained-state measurements depend on learned weights as well as architecture; broader experiments across architectures, initialization schemes, and domains are needed to establish how widely this pattern holds.
The statistic is also sensitive to tokenization, prompt templates, quadruple sampling, hidden-state normalization, and distance choice.
We use Euclidean distances over hidden states and diameter normalization; future work should test cosine distances, centered representations, and alternative normalizers.
Another limitation is the fixed sampling budget for every prompt and layer.
It keeps the experiment comparable, but very short and very long prompts may still have different estimation variance.

\section*{Acknowledgments}

This work was funded by the Frontier Technologies R\&D Program of Jiangsu Province under Grant No.~BF2025012 and by the Beijing Advanced Innovation Center for Future Blockchain and Privacy Computing.

\FloatBarrier

\bibliography{custom}

\end{document}